\documentclass{article} 
\usepackage{nips14submit_e,times}
\usepackage{hyperref}
\usepackage{url}

\usepackage{graphicx}
\usepackage{algorithm}
\usepackage{algorithmicx,algpseudocode}
\usepackage{amsmath}
\usepackage{amssymb}
\usepackage{mathtools}
\usepackage{centernot}
\usepackage{multirow}

\title{A Reduction of the Elastic Net to Support Vector Machines with an Application to GPU Computing}

\author{Quan Zhou* \\ \texttt{zhouq10@mails.tsinghua.edu.cn}\And Wenlin Chen\dag \\ \texttt{wenlinchen@wustl.edu} \And Shiji Song* \\ \texttt{shijis@mail.tsinghua.edu.cn} \And Jacob R. Gardner\dag \\ \texttt{gardner.jake@wustl.edu} \And Kilian Q. Weinberger \dag \\ \texttt{kilian@wustl.edu} \And Yixin Chen \dag \\ \texttt{ychen25@wustl.edu}
\AND Tsinghua University* \\ \texttt{Beijing 100084, China}  \And Washington University in St. Louis\dag \\ \texttt{1 Brookings Drive, MO 63130, USA} }

\nipsfinalcopy 

\begin{document}

\newcommand{\abbrev}[0]{SVEN}
\newcommand{\bx}[0]{\mathbf{x}}
\newcommand{\bxi}[0]{\mathbf{x}^{(i)}}
\newcommand{\bhxi}[0]{\mathbf{\hat x}^{(i)}}
\newcommand{\bX}[0]{\mathbf{X}}
\newcommand{\by}[0]{\mathbf{y}}
\newcommand{\bz}[0]{\mathbf{z}}
\newcommand{\bZ}[0]{\mathbf{Z}}
\newcommand{\bzi}[0]{\mathbf{z}^{(i)}}
\newcommand{\bw}[0]{\mathbf{w}}
\newcommand{\bone}[0]{\mathbf{1}}
\newcommand{\mR}[0]{\mathcal{R}}
\newcommand{\bbeta}[0]{{\boldsymbol \beta}}
\newcommand{\balpha}[0]{{\boldsymbol \alpha}}
\newcommand{\bb}[0]{{\mathbf{b}}}
\newcommand{\indState}{\State\hspace{\algorithmicindent}}

\newcommand{\jake}[1]{\textcolor{blue}{[\textbf{Jake:} #1]}}

\maketitle

\begin{abstract}
The past years have witnessed many dedicated open-source projects that built and maintain implementations of Support Vector Machines (SVM), parallelized for GPU, multi-core CPUs and distributed systems. Up to this point, no comparable effort has been made to parallelize the Elastic Net, despite its popularity in many high impact applications, including genetics, neuroscience and systems biology.
The first contribution in this paper is of theoretical nature. We establish a tight link between two seemingly different algorithms and prove that Elastic Net regression can be reduced to SVM with squared hinge loss classification.
Our second contribution is to derive a practical algorithm based on this reduction.
The reduction enables us to  utilize prior efforts in speeding up and parallelizing SVMs to obtain a highly optimized and parallel solver for the Elastic Net and Lasso.
With a simple wrapper, consisting of only 11 lines of MATLAB$^{\text{\tiny TM}}$ code, we obtain an Elastic Net implementation that naturally utilizes GPU and multi-core CPUs. We demonstrate on twelve real world data sets, that our algorithm yields identical results as the popular (and highly optimized) \emph{glmnet} implementation but is one or several orders of magnitude faster.

\end{abstract}

\section{Introduction}

The Elastic Net~\cite{zou2005regularization} and Lasso as a special case~\cite{hastie2009elements} are arguably two of the most celebrated and widely used feature selection algorithms of the past decade.
The increase in data set sizes has led to a rise in the demand for fast implementations of popular machine learning techniques.
For example in fMRI classification~\cite{pereira2009machine} one can easily obtain data sets with $p\!>\!1,000,000$ voxels. Similarly in genetics~\cite{gustafsson2010gene} genome-wide predictions often have millions of features.

Meanwhile, the increased availability of GPU processing and multi-core CPUs has provided a natural means to scale up algorithms through parallelization. However, some algorithms are easier to parallelize than others. For example, deep (convolutional) neural networks can naturally take advantage of multiple GPUs~\cite{ImageNet}; support vector machines (SVM) have been ported to GPUs~\cite{cotter2011gpu,tyree2014parallel}, multi-core CPUs~\cite{chapelle2007training,tyree2014parallel}  and even distributed systems~\cite{psvm,navia2006distributed}. Although not originally parallelized, \emph{liblinear}~\cite{fan2008liblinear} utilizes clever dual coordinate ascent updates to drastically speed up linear SVMs.

Despite the growing trend, feature selection algorithms have yet to embrace parallel computing platforms in a similar fashion. One of the most popular parallel implementations of Lasso may be \emph{Shotgun}~\cite{kyrola2011parallel}, which however in our experiments does often not outperform the (admittedly highly optimized) single-core Elastic Net implementation \emph{glmnet} by Friedman~\cite{friedman2010regularization}.

This imbalance of parallelization may be in part due to the fact that parallelizing algorithms is hard to do. For example, the GT-SVM implementation of kernel-SVM for GPUs uses handwritten CUDA kernels~\cite{cotter2011gpu} to truly utilize the computing power of modern graphics cards. This is not only highly labor intensive, it also requires constant maintenance as hardware and software standards progress.

In this paper we introduce a different approach to parallelize the Elastic Net, and Lasso as a special case.
Instead of proposing a new hand designed  parallel implementation of the core algorithm, we take inspiration from recent work on machine learning reductions~\cite{jaggi2013equivalence,langford2005relating} and we reduce the Elastic Net  to the squared hinge-loss SVM (without a bias term). We show that this reduction is exact and extremely efficient in practice.
The resulting algorithm, which we refer to as \emph{Support Vector Elastic Net~(\abbrev{})},  naturally takes advantage of the vast existing work on parallel SVMs, immediately providing highly efficient Elastic Net and Lasso implementations on GPUs, multi-core CPUs and distributed systems~\cite{psvm,cotter2011gpu,fan2008liblinear,navia2006distributed,tyree2014parallel}.

We make three main contributions: 1. we prove a theoretical result and derive the non-trivial equivalence between the Elastic Net and SVM with squared hinge loss; 2. we turn this equivalence relationship into a practical algorithm, \abbrev{}, which can solve any Elastic Net or Lasso problem with out-of-the-box (squared hinge-loss) SVM solvers. 3. we evaluate \abbrev{} on twelve real world data sets (eight in the $p\!\gg\! n$ and four in the $n\!\gg\!p$ setting) and show that \abbrev{} is by far the fastest Elastic Net solver to date---outperforming even the most efficient existing  implementation by an order of magnitude across almost all benchmark data sets.

\section{Notation and Background}
\label{sec:back}

Throughout this paper we type vectors in bold ($\bx$), scalars in regular ($C$ or $b$), matrices in capital bold ($\mathbf{X}$). Specific entries in vectors or matrices are scalars and follow the corresponding convention, $i.e.$ the $i^{th}$ dimension of vector $\mathbf{x}$ is $x_i$.
In contrast, depending on the context, $\bxi$ refers to the $i^{th}$ \emph{column} in matrix $\bX$ and $\bx_i$ refers to the transpose of its $i^{th}$ \emph{row}. $\mathbf 1$ is a column vector of all 1. In the remainder of this section we briefly review the Elastic Net and SVM.

\paragraph{Elastic Net.}
In the regression scenario we are provided with a data set $\{(\bx_i,y_i)\}_{i=1}^n$, where each $\bx_i\!\in\!\mR^p$ and the labels are real valued, \emph{i.e.} $y_i\!\in\! \mR$. Let $\by = (y_1, \ldots ,y_n)^\top$ be the response vector and $\bX \!\in\! \mR^{n\times p}$ be the design matrix where the (transposed) $i^{th}$ row of $\bX$ is $\bx_i$. As in \cite{zou2005regularization}, we  assume throughout that the response vector is centered and all features are normalized.

The Elastic Net~\cite{zou2005regularization}  learns a (sparse) linear model to predict $y_i$ from $\bx_i$ by minimizing the squared loss with L2-regularization and an L1-norm constraint,
\begin{equation}
\min_{\bbeta\in \mR^p}\ \|\mathbf{X}\boldsymbol \beta-\mathbf{y}\|_{2}^2+\lambda_2\|\boldsymbol \beta\|_2^2 \qquad
\textrm{such that\ }\ |\boldsymbol \beta|_1\leq t, \ \
\label{eq:enet}
\end{equation}
where $\bbeta=[\beta_1,\ldots,\beta_p]^\top\!\in\!{\mR}^p$ denotes the weight vector, $\lambda_2\!\geq\! 0$ is the L2-regularization constant and $t>0$ the L1-norm budget. In the case where $\lambda_2\!=\!0$, the Elastic Net reduces to the Lasso~\cite{hastie2009elements} as a special case.
The L1 constraint encourages the solution to be sparse.
The L2 regularization coefficient has several desirables effects: 1. it makes the problem strictly convex and therefore yields a unique solution; 2. if features are highly correlated it assigns non-zero weights to all of them (making the solution more stable); 3. if $p\!\gg\! n$ the optimization does not become unstable for large values of $t$.

\paragraph{SVM with squared hinge loss.}
In the classification setting we are given a training dataset $\{(\hat\bx_i,\hat y_i)\}_{i=1}^m$ where $\hat\bx_i\!\in\! \mR^d$ and $\hat y_i\!\in\! \{+1,-1\}$.  The linear SVM with squared hinge loss optimization problem \cite{scholkopf2002learning} learns a separating hyperplane, parameterized by a weight vector $\bw\!\in\! \mR^d$, with the regularized squared hinge loss:
\begin{equation}
\min_{\mathbf w}\ \frac{1}{2}\|\mathbf w\|_{2}^2+ C\sum_{i=1}^{m} \xi_i^2\qquad \textrm{such that\ }~\ \hat{y}_i\mathbf w^\top \hat\bx_i \geq 1-\xi_i~\forall i.
\label{eq:svm1}
\end{equation}
Here, $C\!>\!0$ denotes the regularization constant. Please note that in this paper we do not include any bias term, \emph{i.e.} we assume that the separating hyperplane will pass through the origin.

This problem is often solved in its dual formulation, which due to strong duality is equivalent to solving \eqref{eq:svm1} directly. Without replicating the derivation~\cite{hsieh2008dual,scholkopf2002learning}, we state the dual problem of \eqref{eq:svm1} as:
\begin{equation}
	\min_{\alpha_i\geq 0}  \|\hat\bZ \balpha \|_{2}^2
	+ \frac{1}{2C} \sum_{i=1}^m \alpha_i^2
	-2\sum_{i=1}^m\alpha_i,\label{eq:kernell2svm}
\end{equation}
where $\balpha=\left(\alpha_1,\dots,\alpha_m\right)$ denote the dual variables and $\hat\bZ=\left(\hat y_1 \hat\bx_1,\dots,\hat y_m \hat\bx_m \right)$ is a $d\times m$ matrix, of which the $i^{th}$ column $\bz^{(i)}$ consists of  input $\hat \bx_i$ multiplied by its corresponding label $\hat y_i$, \emph{i.e.} $\bz^{(i)}\!=\!\hat y_i\hat \bx_i$.
The two formulations \eqref{eq:svm1} and \eqref{eq:kernell2svm} are equivalent and the solutions connect via $\bw\!=\!\sum_{i=1}^m \hat y_i\alpha_i \hat\bx_i$.

In~\eqref{eq:kernell2svm} the data is only accessed through  $\hat\bZ^\top\hat \bZ$, which corresponds to the inner-product matrix of the input rescaled by the labels, \emph{i.e.} $[\hat\bZ^\top\hat \bZ]_{ij}=\hat y_i\hat\bx_i^\top\hat\bx_j\hat y_j$. In  scenarios with $d\!\gg\! m$, this matrix can be pre-computed and cached in a kernel matrix in $O(m^2)$ memory and $O(d)$ operations, which makes the remaining running time independent of the dimensionality~\cite{scholkopf2002learning}.
Historically, the dual formulation is most commonly used to achieve non-linear decision boundaries, with the help of the kernel-trick~\cite{scholkopf2002learning}. In our case, however, we will only need the linear setting and restrict the kernel (inner-product matrix) to be linear, too.

Both formulations of the SVM can be solved particularly efficiently on modern hardware with Newton's Method~\cite{chapelle2007training,fan2008liblinear}, which offloads the majority of the computation onto matrix operations and therefore can be vectorized and parallelized to achieve near peak computing performance~\cite{tyree2014parallel}.

In this work, as we do not use the standard SVM with linear hinge loss, we refer to the SVM with squared hinge loss simply as \emph{SVM}.


\section{The Reduction of Elastic Net to SVM }

In this section, we derive the equivalence between Elastic Net and SVM, and reduce problem \eqref{eq:enet} to a specific instance of the SVM optimization problem~\eqref{eq:kernell2svm}.

\renewcommand{\algorithmicrequire}{\textbf{Input:}}
\renewcommand{\algorithmicensure}{\textbf{Output:}}

\paragraph{Reformulation of the Elastic Net.}
We start with the Elastic Net formulation as stated in \eqref{eq:enet}. First, we divide the objective and the constraint by $t$ and substitute in a rescaled weight vector, $\bbeta\coloneqq \frac{1}{t}\bbeta$. This step allows us to absorb the constant $t$ entirely into the objective and rewrite~\eqref{eq:enet}  as
\begin{equation}
\min_{ \bbeta}\ \left\|\bX \bbeta-\frac{1}{t}\by\right\|_{2}^2+\lambda_2\|\boldsymbol \beta\|_2^2\qquad
s.t.\ |\boldsymbol \beta|_1\leq 1. \ \
\label{eq:rescale2}
\end{equation}
To simplify the L1 constraint, we follow~\cite{schmidt2005least} and split $\bbeta$ into two sets of non-negative variables, representing positive components $\bbeta^+\!\geq\! 0$ and negative components $\bbeta^-\!\geq\! 0$, \emph{i.e.\ } $\bbeta=\bbeta^+\!-\!\bbeta^-$.
Then we stack $\bbeta^+$ and $\bbeta^-$ together and form a new weight vector $\hat\bbeta=[\bbeta^+;\bbeta^-]\!\in\! \mR^{2p}_{\geq 0}$.
The regularization term $\|\bbeta\|^2_2$ can be expressed as $\sum_{i=1}^{2p} \hat{\beta_i}^2$, and \eqref{eq:rescale2} can be rewritten as
\begin{equation}
\min_{ \hat\beta_i\geq 0}\ \left\|\left[\bX,-\bX \right] \hat \bbeta-\frac{1}{t}\by\right\|_{2}^2
+\lambda_2\sum_{i=1}^{2p} \hat{\beta_i}^2  \qquad s.t.\ \sum_{i=1}^{2p} \hat\beta_i\leq1.\ \
\label{eq:proof1}
\end{equation}
Here the set $\mR^{2p}_{\geq 0}$ denotes all vectors in $\mR^{2p}$ with all non-negative entries. Please note that, as long as $\lambda_2\!\neq\! 0$, the solution to \eqref{eq:proof1} is unique and satisfies that $\beta_i^+\!=\!0$ or $\beta_i^-\!=\!0$ for all $i$.

Barring the (uninteresting) case with extremely large $t\!\gg\! 0$, the L1-norm constraint in \eqref{eq:enet} will always be tight~\cite{boyd2004convex}, \emph{i.e.} $|\hat \bbeta|\!=\!1$.
(If $t$ is extremely large,~\eqref{eq:enet} is equivalent to ridge regression~\cite{hastie2009elements}, which typically yields completely dense (non-sparse) solutions.)
We can incorporate this equality constraint into~\eqref{eq:proof1} and obtain
\begin{equation}
\min_{\hat\beta_i\geq 0}\ \left\|\left[\bX,-\bX \right] \hat \bbeta-\frac{1}{t}\by\right\|_{2}^2
+\lambda_2\sum_{i=1}^{2p} \hat{\beta_i}^2
\qquad s.t.\ \sum_{i=1}^{2p} \hat\beta_i=1.
\label{eq:equivalence1}
\end{equation}
We construct a matrix $\hat\bZ\!=\![\hat\bX_1,-\hat \bX_2]$ such that $\hat\bZ\hat\bbeta=\!\left[\bX,-\bX\right]\hat\bbeta\!-\!\frac{1}{t}\by$. As $\mathbf{1}^\top\hat\bbeta\!=\!1$, we can expand $\by\!=\!\by\mathbf{1}^\top\hat\bbeta$ and define $\hat\bX_1\!=\!\bX\!-\!\frac{1}{t}\mathbf{y\mathbf 1^\top}$ and $\hat\bX_2\!=\!\bX\!+\!\frac{1}{t}\mathbf{y\mathbf 1^\top}$.
If we substitute $\hat\bZ$ into \eqref{eq:equivalence1} it becomes
\begin{equation}
\min_{\hat\beta_i\geq 0}\ \| \hat\bZ\hat {\bbeta} \|_{2}^2
+\lambda_2\sum_{i=1}^{2p} \hat{\beta_i}^2 \qquad s.t.\ \sum_{i=1}^{2p} \hat\beta_i=1.\ \
\label{eq:equivalence2}
\end{equation}

In the remainder of this section we show that one can obtain the optimal solution $\hat\bbeta^*$ for~\eqref{eq:equivalence2} by carefully constructing a  binary classification data set $\hat \bX,\hat \by$ such that $\hat\bbeta^*\!=\!\balpha^*/|\balpha^*|_1$, where $\boldsymbol \alpha^*$ is the solution for the  SVM dual~\eqref{eq:kernell2svm} for $\hat \bX,\hat \by$.

\paragraph{Data set construction.}
We construct a binary classification data set with $m\!=\!2p$ samples and $d\!=\!n$ features consisting of the columns of $\hat{\bX}\!=\![\hat{\bX}_1, \hat{\bX}_2]$. Let us denote this set as $\{(\hat{\bx}^{(1)},\hat y_1),\dots,(\hat{\bx}^{(2p)},\hat y_{2p})\}$, where each $\hat\bx^{(i)}\!\in\!\mR^{n}$ and $\hat y_1,\dots, \hat y_p\!=\!+1$ and $\hat y_{p+1},\dots, \hat y_{2p}\!=\!-1$. In other words, the columns of $\hat\bX_1$ are of class $+1$ and the columns of $\hat\bX_2$ are of class $-1$.
It is straight-forward to see that for $\hat\bZ\!=\![\hat\bX_1,-\hat \bX_2]$, as used in \eqref{eq:equivalence2}, we have
$\hat \bZ=\left(\hat y_1 \hat\bx_1,\dots,\hat y_m \hat\bx_m \right)$, matching the definition in~\eqref{eq:kernell2svm}. In other words, the solution of \eqref{eq:kernell2svm} with $\hat \bZ$ is the SVM classifier when applied to $\hat \bX,\hat \by$.

\paragraph{Optimal solution.}
Let $\balpha^*$ denote the optimal solution of \eqref{eq:kernell2svm}, when optimized with this matrix $\hat\bZ$ and $C=\frac{1}{2\lambda_2}$.
We will now reshape the SVM optimization problem \eqref{eq:kernell2svm} into the Elastic Net \eqref{eq:equivalence2} without changing the optimal solution, $\balpha^*$ (up to scaling). First, knowing the optimal solution to \eqref{eq:kernell2svm}, we can add the constraint $\sum_{i=1}^{2p}\alpha_i\!=\!|\balpha^*|_1$, which is trivially satisfied at the optimum, $\balpha^*$, and \eqref{eq:kernell2svm} becomes:
\begin{equation}
	\min_{\alpha_i\geq 0}  \|\bZ \balpha \|_{2}^2
	+ \lambda_2 \sum_{i=1}^{2p} \alpha_i^2
	-2\sum_{i=1}^{2p}\alpha_i.
	~~~~~s.t. ~~\sum_{i=1}^{2p}\alpha_i=|\balpha^*|_1.
	\label{eq:equality}
\end{equation}
Because of this equality constraint, the last term in the objective function in~\eqref{eq:equality}, $-2\sum_{i=1}^{2p}\alpha_i\!=\!-2|\balpha^*|_1$, becomes a constant and can be dropped. Removing this constant term does not affect the solution and leads to the following equivalent optimization:
\begin{equation}
	\min_{\alpha_i\geq 0}  \|\bZ \balpha \|_{2}^2
	+ \lambda_2 \sum_{i=1}^{2p} \alpha_i^2.
	~~~~~s.t. ~~\sum_{i=1}^{2p}\alpha_i=|\balpha^*|_1.
	\label{eq:dummy}
\end{equation}
Note that the only difference between \eqref{eq:dummy} and \eqref{eq:equivalence2} is the scale of design variables.
If we divide\footnote{This is not well defined if $|\alpha^*|_1\!=\!0$, which is the degenerate case when the SVM selects no support vectors,
$\balpha\!=\!\mathbf{0}$, and which is not meaningful without bias term.}
the objective by $|\balpha^*|_1^2$ and the constraint by $|\balpha^*|_1$ and introduce a change of variable,
$\hat\beta_i\!=\!\alpha_i/|\balpha^*|_1$ we obtain exactly \eqref{eq:equivalence2} and its optimal solution $\hat\bbeta^*\!=\!\balpha^*/|\balpha^*|_1$.

\paragraph{Implementation details.}
To highlight the fact that this reduction is not just of theoretical value but highly practical, we summarize it in Algorithm~\ref{algo1} in MATLAB$^{\text{\tiny TM}}$ code.\footnote{For improved readability, some variable names are mathematical symbols and would need to be substituted in clear text (\emph{e.g.} $\balpha$ should be \emph{alpha})} We refer to our algorithm as \emph{Support Vector Elastic Net~(\abbrev{})}.
As mentioned in the previous section~\ref{sec:back}, the dual and primal formulations of SVM have different time complexities and we choose the faster one depending on whether $2p\!>\!n$ or vice versa. Line~\ref{alg:wtoalpha} converts the primal variables $\bw$ to the dual solution $\balpha$~\cite{scholkopf2002learning}. 
Many solvers \cite{bottou2010large,chapelle2007training,fan2008liblinear,tyree2014parallel} have been developed for  the linear SVM problem~\eqref{eq:svm1}.
In practice, it is no problem to find an implementation with no bias term. Some implementations we investigate do not use a bias by default (\emph{e.g.} \emph{liblinear}~\cite{fan2008liblinear}) and for others it is trivial to remove~\cite{chapelle2007training}.
In our experiments we use an SVM implementation based on
Chapelle's original exact linear SVM implementation~\cite{chapelle2007training} (which can solve the dual and primal formulation respectively).
The resulting algorithm is exact and uses a combination of conjugate gradient (until the number of potential support vectors is sufficiently small) and Newton steps. The majority of the computation time is spent in the Newton updates.
As pointed out by Tyree et al.~\cite{tyree2014parallel}, the individual Newton steps can be parallelized trivially by using parallel BLAS libraries (which is the default in MATLAB$^{\text{\tiny TM}}$), as it involves almost exclusively matrix operations. We also create a GPU version by casting several key matrices as \emph{gpuArray}, a MATLAB$^{\text{\tiny TM}}$ internal variable type that offloads computation onto the GPU.\footnote{The \emph{gpuArray} was introduced into MATLAB$^{\text{\tiny TM}}$ in 2013. }

\paragraph{Feature selection and Lasso.}
It is worth considering the algorithmic interpretation of this reduction. Each input $\hat{\bx}_i$ in the SVM data set corresponds to a feature of the original Elastic Net problem. Support Vectors correspond to features that are selected, \emph{i.e.} $\beta_i\!\neq\! 0$.
If $\lambda_2\!\rightarrow\! 0$ the Elastic Net becomes LASSO~\cite{hastie2009elements}, which has previously been shown to be equivalent to the hard-margin SVM~\cite{jaggi2013equivalence}. It is reassuring to see that our formulation recovers this relationship as a special case, as $\lambda_2\!\rightarrow\! 0$ implies that $C\!\rightarrow\! \infty$, converting \eqref{eq:svm1} into the hard-margin SVM. (In practice, to avoid numerical problems with very large values of $C$, one can treat this case specially and call a hard-margin SVM implementation in lines~\ref{alg:svmprimal} and \ref{alg:svmdual} of Algorithm~\ref{algo1}.)

\begin{algorithm}[t]         
\caption{MATLAB$^{\text{\tiny TM}}$ implementation of~\abbrev{}.\label{algo1}}             
\label{alg:ensvm}                  
\begin{algorithmic} [1]
\State \textbf{function} $\bbeta$ = \abbrev{}(X, y, t, $\lambda_2$);
\State [n p] = size(X);
\State Xnew = [bsxfun(@minus, X, y./t); bsxfun(@plus, X, y./t)]';
\State Ynew = [ones(p,1); -ones(p,1)];
\If {$2p>n$}
\State $\bw$ = SVMPrimal(Xnew, Ynew, C = 1/(2$\lambda_2$)); \label{alg:svmprimal}
\State $\balpha$  = C * max(1-Ynew.*(Xnew*$\bw$),0);\label{alg:wtoalpha}
\Else
\State $\balpha$ = SVMDual(Xnew, Ynew, C = 1/(2$\lambda_2$)); \label{alg:svmdual}
\EndIf
\State $\bbeta$ = t * ($\balpha$(1:p) - $\balpha$(p+1:2p)) / sum($\balpha$);
\end{algorithmic}
\end{algorithm}



\paragraph{Time complexity.} The construction of the input only requires $O(np)$ operations and the majority of the running time will, in all cases, be spent in the SVM solver. As a result, the running time of our algorithm has great flexibility, and for any dataset with $n$ inputs and $p$ dimensions, we can choose an SVM implementation with a running time that is advantageous for that dataset. Chapelle's MATLAB$^{\text{\tiny TM}}$ implementation can scale in the worst case either $O(n^3)$ (primal mode) or $O(p^3)$ (dual mode)~\cite{chapelle2007training}.\footnote{As it is slightly confusing it is worth re-emphasizing that if the original data has $n$ samples with $p$ dimensions, the constructed SVM problem has $2p$ samples with $n$ dimensions. } In the typical case the running times are known to be much better. Especially for the dual formulation we can in practice achieve a running time much better than $O(p^3)$, as the worst case assumes that all points are support vectors. In the Elastic Net setting, this would correspond to all features being kept. A more realistic practical running time is on the order of $O(\min(p,n)^2)$, depending on the number of features selected (as regulated by $t$).

SVM implementations with other running times can easily be adapted to our setting, for example \cite{joachims2006training} would allow training in time $O(np)$ and recent work even suggests solvers with sub-linear time complexity~\cite{hazan2011beating} (although the solution might be insufficiently exact for our purposes in practice).


\section{Related Work}
The Elastic Net has been widely deployed in many machine learning applications, but only little effort has been made towards efficient parallelization. The coordinate gradient descent algorithm has become the dominating strategy for the optimization.

The state-of-the-art single-core implementation for solving the Elastic Net problem is the \emph{glmnet} package developed by Friedman~\cite{friedman2010regularization}. Mostly written in Fortran language, \emph{glmnet} adopts the coordinate gradient descent strategy and is highly optimized. As far as we know, it is the fastest  off-the-shelf solver for the Elastic Net.
Due to its inherent sequential nature, the coordinate descent algorithm is extremely hard to parallelize. The \emph{Shotgun} algorithm proposed by \cite{kyrola2011parallel} is among the first to parallelize coordinate descent for Lasso.
This implementation can run on extremely sparse large scale datasets that other software, including \emph{glmnet}, cannot run on due to memory constraints.
The \emph{L1\_LS} algorithm proposed by \cite{kim2007interior} transforms the Lasso to its dual form directly and uses a log-barrier interior point method for optimization.
The optimization is based on using the Preconditioned Conjugate Gradient (PCG) method to solve Newton steps which is suitable for sparse large scale compressed sensing problems.

On the SVM side, one of the most popular and user-friendly implementations is the \emph{libsvm} library~\cite{chang2011libsvm}. However, it is optimized to solve kernel SVM problems using sequential minimal optimization (SMO)~\cite{platt1998sequential}, which is not efficient for the specific case of linear SVM. The \emph{liblinear} library \cite{fan2008liblinear} is specially tailored for linear SVMs, including the squared hinge loss version. However, we did find that on modern multi-core platforms (with and without GPU acceleration) algorithms that actively seek updates through matrix operations~\cite{tyree2014parallel} tend to be substantially faster (in both settings, $p\!\gg\! n$ and $n\!\gg\! p$).

Our work is inspired by a recent theoretical contribution, Jaggi 2013~\cite{jaggi2013equivalence}, which reveals the close relation between Lasso and hard-margin SVMs. We extend this line of work and prove a non-trivial equivalence between the Elastic Net and the soft-margin SVM and we derive a practical algorithm,  which we validate experimentally.

\begin{figure}[t]
\begin{center}
\includegraphics[width=0.8\textwidth]{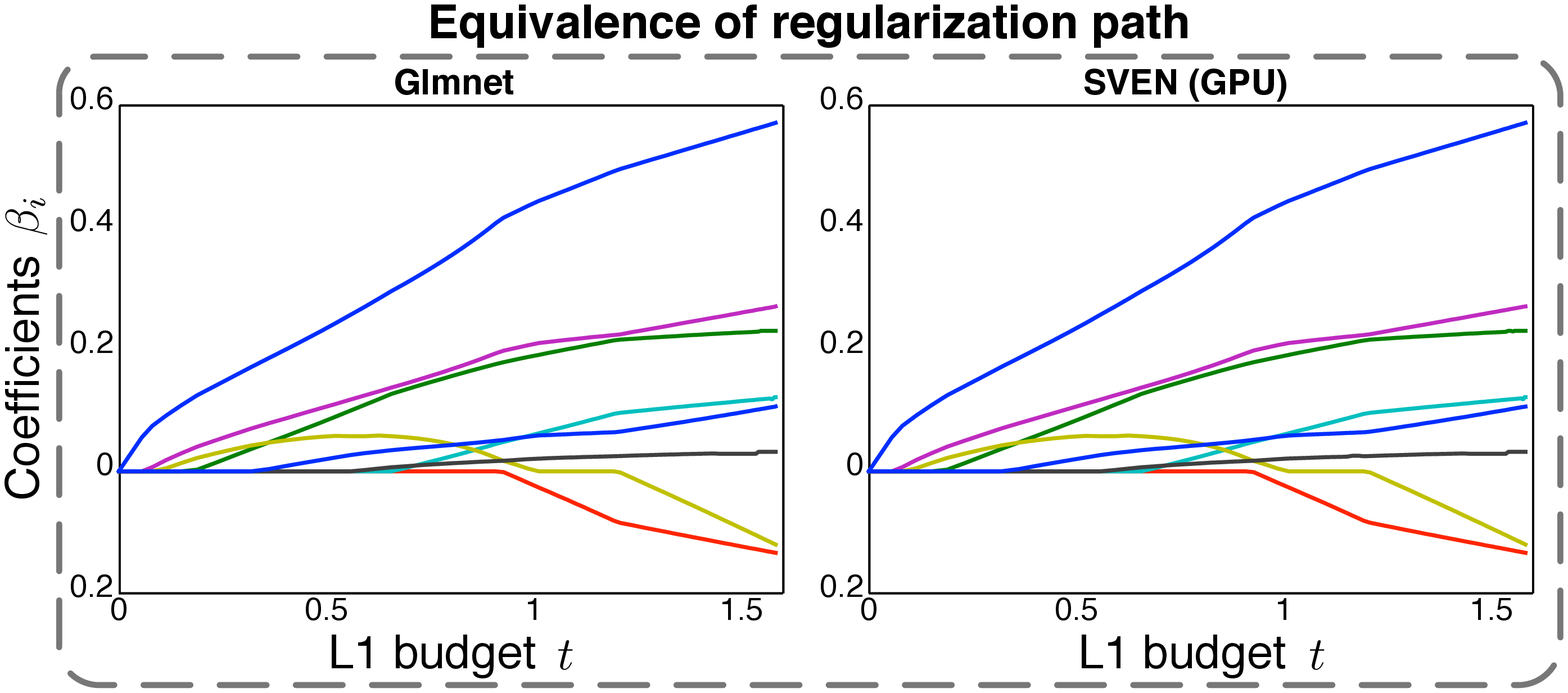}
\end{center}
\vspace{-3ex}
\caption{The regularization paths of \emph{glmnet} (left) and \abbrev{}~(GPU) (right) on the \emph{prostate} dataset. Each line corresponds to the value of $\beta_i^*$ as a function of the L1 budget $t$. The two algorithms match exactly for all values of $t$. }
\label{fig:solpath}
\end{figure}

\section{Experimental Results and Conclusion}
In this section, we conduct extensive experiments to evaluate \abbrev{} on twelve real world data sets. We first provide a brief description of the experimental setup and the data sets, then we investigate the two common scenarios $p\!\gg\! n$ and $n\!\gg\!p$ separately. For full replicability of the experiments, our source code and links to data sets are available online at {\small\url{http://anonymized}}.

\paragraph{Experimental Setting.}
We test our method on GPU and (multi-core) CPU under the names of \abbrev{} (GPU) and \abbrev{} (CPU), respectively.
For comparison, we have a single-threaded CPU baseline method: \emph{glmnet} \cite{friedman2010regularization}, a popular and highly optimized Elastic Net software package.
On multi-cores we evaluate two parallelized Lasso implementations. The \emph{Shotgun} algorithm by Bradley et al.~\cite{kyrola2011parallel} parallelizes coordinate gradient descent. Finally we also compare against \emph{L1\_LS}, a parallel MATLAB solver (for Lasso) implemented by Kim et al.~\cite{kim2007interior}.
All the experiments were performed on an off-the-shelve desktop with two 8-core Intel(R) Xeon(R) processors of 2.67 GHz and $96GB$ of RAM. The attached NVIDIA GTX TITAN graphics card contains 2688 cores and 6 GB of global memory.


\paragraph{Regularization path.} On all data sets we compare $40$ different settings for $\lambda_2$ and $t$. We obtain
these by first solving for the full solution path with \emph{glmnet}. The \emph{glmnet} implementation enforces the L1
budget not as a constraint, but as an L1-regularization penalty with a regularization constant $\lambda_1$. We obtain
the solution path by slowly decreasing $\lambda\!=\!\lambda_1\!+\!\lambda_2$. We sub-sample $40$ evenly spaced settings along this path that lead to solutions with distinct number of selected features. If the \emph{glmnet} solution for a particular
parameter setting is $\beta^*$ we obtain $t$ by computing $t=|\beta^*|_1$. This procedure provides us with $40$
parameter pairs $(\lambda_2,t)$ for each data set on which we compare all algorithms. (For the pure Lasso
implementations, \emph{shotgun} and \emph{L1\_LS},
we set $\lambda_2=0$.)


\paragraph{Correctness.} Throughout all experiments and all settings of $\lambda_2$ and $t$ we find that \emph{glmnet} and \abbrev{} obtain identical results up to the tolerance level.
To illustrate the equivalence, Figure~\ref{fig:solpath} shows the regularization path of \abbrev{} (GPU) and \emph{glmnet} on the \emph{prostate cancer} data used in \cite{zou2005regularization}. As mentioned in the previous paragraph, we obtain the original solution path from \emph{glmnet}  and evaluate \abbrev{}~(GPU) on these parameter settings.
The data has eight clinical features (\emph{e.g.} log(cancer volume), log(prostate weight)) and the response is the logarithm of prostate-specific antigen (lpsa).
Each line in Figure~\ref{fig:solpath} corresponds to the $\beta_i^*$ value of some feature $i\!=\!1,\dots,8$ as a function of the L1 budget $t$. The graph indicates that the two algorithms lead to exactly matching regularization paths as the budget $t$ increases.

\begin{figure}[t]
\begin{center}
\includegraphics[width=\textwidth]{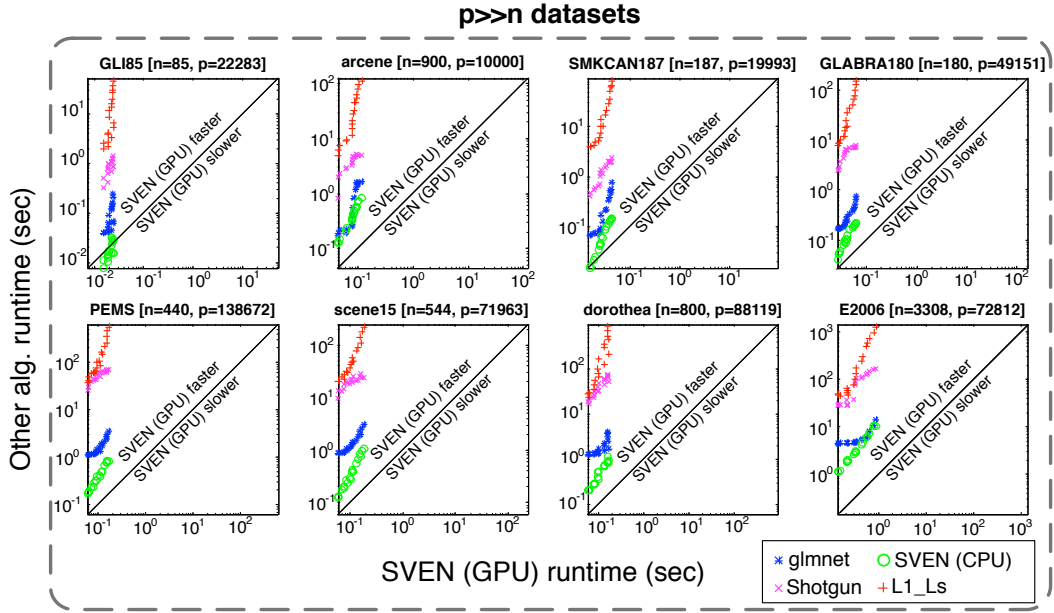}
\vspace{-5ex}
\end{center}
\caption{Training time comparison of various algorithms in $p\!\gg\!n$ scenarios. Each marker compares an algorithm with \abbrev{}~(GPU) on one (out of eight) datasets and one parameter setting. The X,Y-axes denote the running time of \abbrev{}~(GPU) and that particular algorithm on the same problem, respectively. All markers are above the diagonal line (except \abbrev{}~(CPU) for GLI-85),  indicating that \abbrev{}~(GPU) is faster than all baselines in all cases.}
\label{fig:speedup1}
\end{figure}

\paragraph{Data sets with $p\!\gg\! n$.}
The $p\! \gg\! n$ scenario may be the most common application setting for the Elastic Net and Lasso and there is an abundance of real world data sets. We evaluate all methods on the following eight of them: GLI-85, a dataset that screens a large number of diffuse infiltrating gliomas through transcriptional profiling; SMK-CAN-187, a  gene expression dataset from smokers w/o lung cancer; GLA-BRA-180, a  dataset concerning analysis of gliomas of different grades; Arcene, a  dataset from the NIPS 2003 feature selection contest, whose task is to distinguish cancer versus normal patterns from mass spectrometric data; Dorothea, a sparse dataset from the NIPS 2003 feature selection contest, whose task is to predict which compounds bind to Thrombin.\footnote{We removed features with all-zero values across all inputs.} Scene15, a  scene recognition data set \cite{lazebnik2006beyond,xu2012greedy} where we use the binary class 6 and 7 for feature selection; PEMS \cite{Bache+Lichman:2013}, a  dataset that describes the occupancy rate, between 0 and 1, of different car lanes of San Francisco bay area freeways. E2006-tfidf, a sparse dataset whose task is to predict risk from financial reports based on TF-IDF feature representation.\footnote{Here, we reduce the training set size by subsampling to match the size of the test set, $n\!=\!3308$.}

\paragraph{Evaluation ($p\!\gg\! n$).}
Figure \ref{fig:speedup1} depicts training time comparisons of the three baseline algorithms and \abbrev{}~(CPU) on the eight datasets with \abbrev{}~(GPU).  Each marker corresponds to a comparison of one algorithm and \abbrev{}~(GPU) in one particular setting along the regularization path.
 It's  y-coordinate corresponds to the training time required for the corresponding algorithm and
its x-coordinate corresponds to the training time required for \abbrev{}~(GPU) with the exact same L1-norm budget and $\lambda_2$ value. All markers above the diagonals corresponds to runs where \abbrev{}~(GPU) is faster, and all markers below the diagonal corresponds to runs where \abbrev{}~(GPU) is slower.

We observe several general trends: 1. Across \emph{all} eight data sets \abbrev{}~(GPU) always outperforms \emph{all} baselines. The only markers below the diagonal are from \abbrev{}~(CPU) on the GLI-85 data set, which is the smallest and where the transfer time for the GPU is not offset by the gains in more parallel computation. 2. Even \abbrev{}~(CPU) outperforms or matches the performance of the fastest baseline across all data sets. 3. As the L1-budget $t$ increases, the training time increase for all algorithms, but much more strongly for the baselines than for \abbrev{}~(GPU). This can be observed by the fact that the markers of one color (\emph{i.e. } one algorithm) follow approximately lines with much steeper slope than the diagonal.

\begin{figure}[t]
\begin{center}
\includegraphics[width=\textwidth]{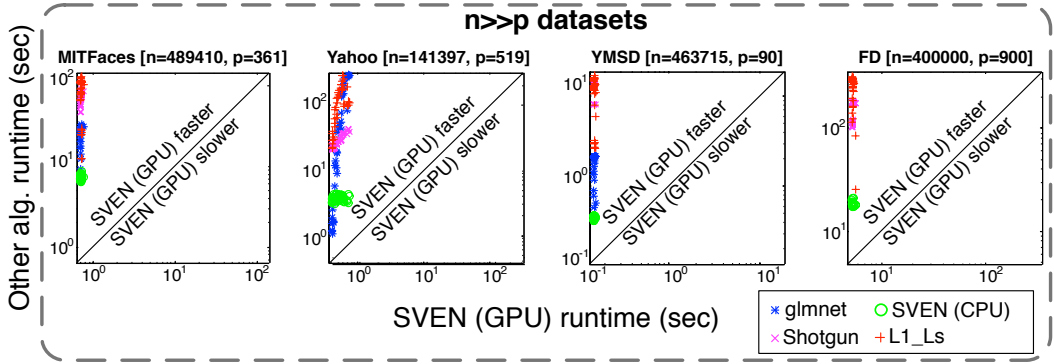}
\end{center}
\vspace{-3ex}
\caption{Training time comparison of various algorithms in $n\!\gg\!p$ scenarios. Each marker compares an algorithm with \abbrev{}~(GPU) on one (out of four) datasets and one parameter setting. The X,Y-axes denote the running time of \abbrev{}~(GPU) and that particular algorithm on the same problem, respectively.
All markers are above the diagonal line,  as \abbrev{}~(GPU) is faster in all cases.}
\vspace{-1ex}
\label{fig:speedup2}
\end{figure}

\paragraph{Data sets with $n\!\gg\! p$.}
For the $n \!\gg\! p$ setting, we evaluate all algorithms on four additional datasets. MITFaces, a facial recognition dataset; the Yahoo learning to rank dataset, a dataset concerning the ranking of webpages in response to a search query; YearPredictionMSD (YMSD), a dataset of songs with the goal to predict the release year of a song from audio features; and FD, another face detection dataset.

\paragraph{Evaluation ($n\!\gg\!p$).} A comparison to all methods on all four datasets can be found in Figure~\ref{fig:speedup2}. The speedups of \abbrev{}~(GPU) are even more pronounced in this setting. The training time of \abbrev{}~(GPU) is completely dominated by the kernel computation and therefore almost identical for all values of $t$ and $\lambda_2$. Consequently all markers follow vertical lines in all plots. The massive speedup of up to two orders of magnitude obtained by \abbrev{}~(GPU) over the baseline methods squashes all markers at the very left most part of the plot.
\emph{glmnet} failed to complete the FD dataset due to memory constraints and therefore we evaluated on the $\lambda_2$  and $t$ values along the solution path from the other face recognition data set, MITFaces.
As in the $p \!\gg\! n$ case, all solutions returned by both versions of \abbrev{} match those of \emph{glmnet} exactly.


\subsection{Discussion}

The use of algorithmic reduction to obtain parallelization and improved scalability has several highly compelling advantages: 1. no new learning algorithm has to be implemented and optimized by hand (besides the small transformation code); 2.  the burden of code maintenance reduces to the single highly optimized (SVM) algorithm; 3. the implementation is very reliable from the start as almost the entire execution time is spent in a well established and tested implementation; 4. and finally, target algorithm may lend itself much more naturally to parallelization.
In our case, the squared hinge-loss SVM formulation can be solved almost entirely with large matrix operations, which are already parallelized (and maintained) by high-performance experts through BLAS libraries (\emph{e.g.} CUBLAS for NVIDIA GPUs {\small \emph{\url{http://tinyurl.com/cublas.}}}).
We hope that this paper will benefit the community in at least two ways: Practitioners will obtain a new stable and blazingly fast implementation of Elastic Net and Lasso; and machine learning researchers might become inspired to identify and derive different algorithmic reductions to facilitate similar performance improvements with other learning algorithms.


\paragraph{Acknowledgements.} QZ and SS are supported by Key Technologies Program of China grant 2012BAF01B03, Research Fund for the Doctoral Program of Higher Education 20130002130010, 20120002110035 and NSFC grant 61273233. KQW, JRG, YC, and WC are supported by NIH grant U01 1U01NS073457-01 and NSF grants 1149882, 1137211, CNS-1017701, CCF-1215302, IIS-1343896.  Computations were performed via the Washington University Center for High Performance Computing, partially through grant NCRR 1S10RR022984-01A1. The authors thank Martin Jaggi for clarifying discussions and suggestions.

\small
\bibliography{nips14_sven.bbl}
\bibliographystyle{plain}

\end{document}